\documentclass[11pt,a4paper]{article}
\usepackage{authblk}
\usepackage[hyperref]{naaclhlt2018}
\usepackage{times}
\usepackage{latexsym}
\usepackage{graphicx}
\usepackage{xcolor}
\usepackage{amsfonts}
\usepackage{amsmath}
\usepackage{booktabs}
\usepackage{multirow}
\usepackage{tikz}
\usepackage{tikz-qtree}
\usepackage{subcaption}

\usepackage{url}

\aclfinalcopy 
\def\aclpaperid{16} 

\DeclareMathOperator*{\argmin}{arg\,min}

\title{Embedding Text in Hyperbolic Spaces}

\author[1]{Bhuwan Dhingra\thanks{\enskip Work done while interning at Google Brain.} }
\author[2]{Christopher J. Shallue}
\author[2]{Mohammad Norouzi}
\author[2]{\\Andrew M. Dai}
\author[2]{George E. Dahl}
\affil[1]{Carnegie Mellon University}
\affil[2]{Google Brain}
\affil[ ]{{\tt bdhingra@cs.cmu.edu, \{shallue, mnorouzi, adai, gdahl\}@google.com}}

\date{}

\begin{document}
\maketitle
\begin{abstract}
Natural language text exhibits hierarchical structure in a variety of respects. Ideally, we could incorporate our prior knowledge of this hierarchical structure into unsupervised learning algorithms that work on text data. Recent work by \citet{nickel2017poincar} proposed using hyperbolic instead of Euclidean embedding spaces to represent hierarchical data and demonstrated encouraging results when embedding graphs. In this work, we extend their method with a re-parameterization technique that allows us to learn hyperbolic embeddings of arbitrarily parameterized objects. We apply this framework to learn word and sentence embeddings in hyperbolic space in an unsupervised manner from text corpora. The resulting embeddings seem to encode certain intuitive notions of hierarchy, such as word-context frequency and phrase constituency. However, the implicit continuous hierarchy in the learned hyperbolic space makes interrogating the model's learned hierarchies more difficult than for models that learn explicit edges between items. The learned hyperbolic embeddings show improvements over Euclidean embeddings in some -- but not all -- downstream tasks, suggesting that hierarchical organization is more useful for some tasks than others.
\end{abstract}

\section{Introduction}
\label{sec:intro}

Many real-world datasets exhibit hierarchical structure, either explicitly in ontologies like WordNet, or implicitly in social networks \citep{adcock2013tree} and natural language sentences \citep{everaert2015structures}. When learning representations of such datasets, hyperbolic spaces have recently been advocated as alternatives to the standard Euclidean spaces in order to better represent the hierarchical structure \citep{nickel2017poincar,chamberlain2017neural}. Hyperbolic spaces are non-Euclidean geometric spaces that naturally represent hierarchical relationships; for example, they can be viewed as continuous versions of trees \citep{krioukov2010hyperbolic}. Indeed, \citet{nickel2017poincar} showed improved reconstruction error and link prediction when embedding WordNet and scientific collaboration networks into a hyperbolic space of small dimension compared to a Euclidean space of much larger dimension.

\begin{figure}[t!]
\centering
\includegraphics[width=\linewidth]{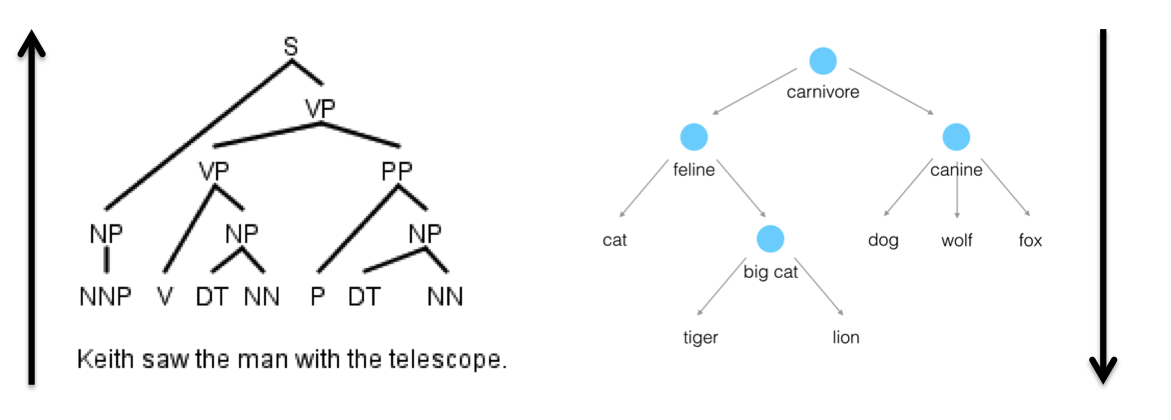}
\caption{Two examples of hierarchical structure in natural language. \textbf{Left:} A constituent parse tree. \textbf{Right:} A fragment of WordNet. Arrows represent the direction in which the nodes become semantically more specific.}
\label{fig:parse-tree}
\end{figure}

In this work, we explore the use of hyperbolic spaces for embedding natural language data, which has natural hierarchical structure in terms of \textit{specificity}. For example, sub-phrases in a sentence can be arranged into a consituency-based parse tree where each node is semantically more specific than its parent (Figure~\ref{fig:parse-tree} left). This hierarchical structure is not usually annotated in text corpora. Instead, we hypothesize that this structure is implicitly encoded in the range of natural language contexts in which a concept appears: semantically general concepts will occur in a wider range of contexts than semantically specific ones. We use this intuition to formulate unsupervised objectives for learning hyperbolic embeddings of text objects. By contrast, \citet{nickel2017poincar} only embedded graphs with an explicit hierarchical structure.

Further, \citet{nickel2017poincar} only considered the non-parametric case where each object to be embedded is assigned its representation from a lookup table\footnote{Note that the term ``non-parametric'' has a different meaning here than in the case of Bayesian non-parametric statistics. Here it refers to the fact that the embeddings are not output by a parameterized function.}. This approach is impractical for embedding natural language because there are too many sentences and phrases for such a table to fit in memory. For natural language, we must adopt a parametric approach where we learn the parameters $\theta$ of an encoder function $f_\theta$ that maps sequences of text to their embeddings. When training their non-parametric model, \citet{nickel2017poincar} relied on a projection step to keep their embeddings within their model of hyperbolic space. Specifically, they embedded their data in the Poincar\'e ball model of hyperbolic space, which consists of points in the unit ball $\mathcal{B}^d = \{\mathbf{x} \in \mathbb{R}^d: \|\mathbf{x}\|<1\}$, but their Reimannian gradient-descent algorithm was not guaranteed to keep their embeddings within the unit ball. To address this issue, they applied a projection step after each gradient step to force the embeddings back into the unit ball, but this projection is not possible when the representations are the output of an encoder $f_\theta$. 

Our main contribution is to propose a simpler parametrization of hyperbolic embeddings that allows us to train parametric encoders. We avoid the need for a projection step by separately parameterizing the direction and norm of each embedding and applying a sigmoid activation function to the norm. This ensures that embeddings always satisfy $\|\mathbf{e}\| < 1$ (as required by the Poincar\'e ball model of hyperbolic space), even after arbitrary gradient steps. Once the embeddings are constrained in this way, all that is needed to induce hyperbolic embeddings is an appropriate distance metric (see Equation~\ref{eq:hyperbolic}) in the loss function in place of the commonly used  Euclidean or cosine distance metrics. In addition to allowing parametric encoders, this parameterization has an added benefit that instead of Riemannian-SGD \citep[as used in][]{nickel2017poincar}, we can use any of the popular optimization methods in deep learning, such as Adam \citep{kingma2014adam}. We show that re-parameterizing in this manner leads to comparable reconstruction error to the method of \citet{nickel2017poincar} when learning non-parametric embeddings of WordNet.

We test our framework by learning unsupervised embeddings for two types of natural language data. First, we embed a graph of word co-occurrences extracted from a large text corpus. The resulting embeddings are hierarchically organized such that words occurring in many contexts are placed near the origin and words occurring in few contexts are placed near the boundary of the space. Using these embeddings, we see improved performance on a lexical entailment task, which supports our hypothesis that co-occurrence frequency is indicative of semantic specificity. However, this improvement comes at the cost of worse performance on a word similarity task. In the second experiment, we learn embeddings of sentences (and sub-sentence sequences) by applying the hyperbolic metric to a modified version of the Skip-Thoughts model \citep{kiros2015skip} that uses embeddings to predict local context in a text corpus. Since most sentences are unique, there is no clear notion of co-occurrence frequency in this case. However, we find a high correlation ($0.67$) between the norms of embedded constituent phrases from Penn Treebank \citep{marcus1993building} and the height at which those phrases occur in their parse trees. We conclude that hyperbolic sentence embeddings encode some of the hierarchical structure represented by parse trees, without being trained to do so. However, experiments on downstream tasks do not show consistent improvements over baseline Euclidean embeddings.

\section{Background -- Poincar\'e Embeddings}

In this section we give an overview of the Poincar\'e embeddings method from \citet{nickel2017poincar}. A similar formulation was also presented in \citet{chamberlain2017neural}.

A hyperbolic space is a non-Euclidean geometric space obtained by replacing Euclid's parallel postulate with an alternative axiom.
The parallel postulate asserts that for every line $L$ and point $P$ not on $L$, there is a unique line co-planar with $P$ and $L$ that passes through $P$ and does not intersect $L$. In hyperbolic geometry, this axiom is replaced with the assertion that there are at least two such lines passing through $P$ that do not intersect $L$ (from which one can prove that there must be infinitely many such lines). 
In this geometry, some familiar properties of Euclidean space no longer hold; for example, the sum of interior angles in a triangle is less than 180 degrees. Like Euclidean geometry, hyperbolic geometry can be extended to $d$-dimensions. $d$-dimensional hyperbolic space is unique up to a ``curvature'' constant $K$$<$$0$ that sets the length scale. Without loss of generality we assume $K$$=$$-1$.

In hyperbolic space, circle circumference ($2\pi \sinh r$) and disc area ($2\pi (\cosh r -1)$) grow exponentially with radius, as opposed to Euclidean space where they only grow linearly and quadratically. This makes it particularly efficient to embed hierarchical structures like trees, where the number of nodes grows exponentially with depth \citep{krioukov2010hyperbolic}. We hope that such embeddings will simultaneously capture both the similarity between objects (in their distances), and their relative depths in the hierarchy (in their norms).

There are several ways to model hyperbolic space within the more familiar Euclidean space. Of these, the Poincar\'e ball model is most suited for use with neural networks because its distance function is differentiable and it imposes a relatively simple constraint on the representations \citep{nickel2017poincar}. Specifically, the Poincar\'e ball model consists of points within the unit ball
$\mathcal{B}^d$, 
in which the distance between two points $\mathbf{u}, \mathbf{v} \in \mathcal{B}^d$ is
\begin{equation}
\small
\label{eq:hyperbolic}
d(\mathbf{u}, \mathbf{v}) = \cosh^{-1} \left(1 + 2 \frac{\|\mathbf{u}-\mathbf{v}\|^2}{(1-\|\mathbf{u}\|^2)(1-\|\mathbf{v}\|^2)}\right).
\end{equation}

Notice that, as $\|\mathbf{u}\|$ approaches 1, its distance to almost all other points increases exponentially. Hence, an effective tree representation will place root nodes near the origin and leaf nodes near the boundary to ensure that root nodes are relatively close to all points while leaf nodes are relatively distant from most other leaf nodes.

In order to learn representations $\Theta=\{\mathbf{\theta_i}\}_{i=1}^n$ for a set of objects $\mathcal{S}=\{s_i\}_{i=1}^n$, we must define a loss function $\mathcal{L}(\Theta, d)$ that minimizes the hyperbolic distance between embeddings of similar objects and maximizes the hyperbolic distance between embeddings of different objects. Then we can solve the following optimization problem
\begin{equation}
\label{eq:hyperbolic_opt}
\hat{\Theta} = \argmin_{\Theta} \mathcal{L}(\Theta, d) \quad \text{s.t.} \quad \|\mathbf{\theta_i}\| < 1 \enskip \forall \mathbf{\theta_i} \in \Theta
\end{equation}

\citet{nickel2017poincar} use Riemannian-SGD to optimize Equation \ref{eq:hyperbolic_opt}. This involves computing the Riemannian gradient (which is a scaled version of the Euclidean gradient) with respect to the loss, performing a gradient-descent step, and projecting any embeddings that move out of $\mathcal{B}^d$ back within its boundary. In the following section, we propose a re-parametrization of Poincar\'e embeddings that removes the need for the projection step and allows the use of any of the popular optimization techniques in deep learning, such as Adam.

\section{Parametric Poincar\'e Embeddings}\label{sec:reparametrize}

Our goal is to learn a function $f: \mathcal{S} \to \mathcal{B}^d$ that maps objects from a set $\mathcal{S}$ to the Poincar\'e ball $\mathcal{B}^d$.
However, the encoders typically used in deep learning, such as LSTMs, GRUs, and feed-forward networks, may produce representations in  arbitrary subspaces of $\mathbb{R}^{d^{\prime}}$. We introduce a re-parameterization technique that maps $\mathbb{R}^{d^{\prime}}$ to $\mathcal{B}^{d}$ and can be used on top of any existing encoder. Let $\mathbf{e}(s) \in \mathbb{R}^{d^{\prime}}$ denote the output of the original encoder for a given $s \in \mathcal{S}$. The re-parameterization involves computing a direction vector $\mathbf{v}$ and a norm magnitude $p$ from $\mathbf{e}(s)$ as follows:
\begin{align*}
\mathbf{\bar{v}} = \phi_{dir}(\mathbf{e}(s)), &\quad \mathbf{v} = \frac{\mathbf{\bar{v}}}{\|\mathbf{\bar{v}}\|}, \\
\bar{p} = \phi_{norm}(\mathbf{e}(s)), &\quad p = \sigma (\bar{p}),
\end{align*}
where $\phi_{dir}: \mathbb{R}^{d^{\prime}} \to \mathbb{R}^{d}$, $\phi_{norm}: \mathbb{R}^{d^{\prime}} \to \mathbb{R}$ can be arbitrary parametric functions, whose parameters will be optimized during training, and $\sigma$ is the sigmoid function that ensures the resulting norm $p \in (0,1)$. We will introduce specific instantiations of $\phi_{dir}$ and $\phi_{norm}$ in the subsections below.  The re-parameterized embedding is defined as $\mathbf{\theta} = p\mathbf{v}$, which lies in $\mathcal{B}^d$.

Let $w$ denote the model parameters in $\mathbf{e}(s)$, $\phi_{dir}$, and $\phi_{norm}$. We wish to optimize a loss function $\mathcal{L}(w, d)$ that minimizes the hyperbolic distance $d$ between embeddings of similar objects and maximizes the hyperbolic distance between embeddings of dissimilar objects. Since the embeddings $\mathbf{\theta} $ are guaranteed to lie in $\mathcal{B}^d$, we can use any of the optimization methods popular in deep learning -- we use Adam \citep{kingma2014adam}.

Next we discuss specific instantiations of encoders, re-parameterization  functions and loss functions for three types of problems.  

\subsection{Non-Parametric Supervised Embeddings}\label{sec:nonparam_sup}
First, we test our re-parametrization by embedding the WordNet hierarchy with a non-parametric encoder -- the same task considered by \citet{nickel2017poincar}. The dataset is represented by a set of tuples $\mathcal{D} = \{(u,v)\}$, where each pair $(u,v)$ denotes that $u$ is a parent of $v$. Since $u$ and $v$ come from a fixed vocabulary of objects, we use a lookup table $L$ as the base encoder, i.e, $\mathbf{e}(u) = L(u) \in \mathbb{R}^{d+1}$.
We set $\phi_{dir} = \mathbf{x}_{1:d}$ and $\phi_{norm} = \mathbf{x}_{d+1}$ to be slicing functions that extract the first $d$ and the $(d+1)$-th dimensions respectively.

We use the same loss function as \citet{nickel2017poincar}, which uses negative samples $\mathcal{N}(u) = \{v: (u,v) \notin \mathcal{D}, v \neq u\}$ to maximize distance between embeddings of unrelated objects:
\begin{equation*}
\small
\mathcal{L}(w,d) = - \sum_{(u,v) \in \mathcal{D}} \log \frac{e^{-d(u,v)}}{\sum_{v' \in \mathcal{N}(u) \cup \{v\}} e^{-d(u,v')}}.
\end{equation*}

Note that this loss function makes no use of the direction of the edge $(u,v)$, because $d(u,v)$ is symmetric. Nevertheless, we expect it to recover the hierarchical structure of $\mathcal{D}$.

\subsection{Non-Parametric Unsupervised Word Embeddings}\label{sec:nonparam_unsup}
Next, we consider the problem of embedding words from a vocabulary $\mathcal{S}_{\mathcal{V}}$ given a text corpus $\mathcal{T} = (w_1, \ldots, w_{|\mathcal{T}|})$, where $w_i \in \mathcal{S}_{\mathcal{V}}$.

Traditional unsupervised methods, like word2vec \citep{mikolov2013distributed} and GloVe \citep{pennington2014glove}, are optimized for preserving semantic similarity: the embeddings of similar words should be close, and the embeddings of semantically different words should be distant. Remarkably, these unsupervised embeddings also exhibit structural regularities, such as vector offsets corresponding to male-to-female or singular-to-plural transformations \citep{mikolov2013linguistic,mikolov2013distributed}. In this work, by embedding in hyperbolic space, we hope to encode both semantic similarity (in the hyperbolic distances between embeddings) and semantic specificity (in the hyperbolic norms of embeddings). Our hypothesis is that words denoting more general concepts will appear in varied contexts and hence will be placed closer to the origin -- similar to how nodes close to the root in WordNet are placed close to the origin in \citet{nickel2017poincar}. Tasks that rely on a hierarchical relationship between words might benefit from embeddings with these properties.

The idea of using specialized vector space models for encoding various lexical relations was previously explored by \citet{henderson2016vector}. While they looked exclusively at the entailment relation, the notion we study here is that of \textit{semantic specificity}, which is more general but also difficult to define formally. One example is that ``musician'' is related to ``music'' and more specific than it, but not necessarily entailed by it.

Both word2vec and GloVe embed words using co-occurrences of pairs of words occur within a fixed window size in $\mathcal{T}$. Here, we construct a co-occurrence graph $\mathcal{G}=\{(w,v)\}$ that consists of all pairs of words that occur within a fixed window of each other. Certain pairs co-occur more frequently than others, and we preserve this information by allowing repeated edges in $\mathcal{G}$: each pair $(w,v)$ occurs $f^c$ times in $\mathcal{G}$, where $f$ is the frequency of that pair in $\mathcal{T}$ and $c < 1$ is a downsampling constant. We embed $\mathcal{G}$ in the Poincar\'e ball in the manner described in Section~\ref{sec:nonparam_sup}.

\subsection{Parametric Unsupervised Sentence Embeddings}\label{sec:param_sentences}
Finally, we consider embedding longer units of text such as sentences and phrases. We denote the set of all multi-word expressions of interest as $\mathcal{S}_{\mathcal{Z}}$. Our goal is to learn an encoder function $f: \mathcal{S}_{\mathcal{Z}} \to \mathcal{B}^d$ in an unsupervised manner from a text corpus $\mathcal{T} = (s_1, \ldots, s_{|\mathcal{T}|})$, where $s_i \in \mathcal{S}_{\mathcal{Z}}$.

Sentence embeddings are motivated by the phenomenal success of word embeddings as general purpose feature representations for a variety of downstream tasks. The desiderata of multi-word embeddings are similar to those of word embeddings: semantically similar units should be close to each other in embedding space, and complex semantic properties should map to geometric properties in the embedding space. Our hypothesis is that embedding multi-word units in hyperbolic space will capture the hierarchical structure of specificity of the meanings of these units.


We start with Skip-Thoughts \citep{kiros2015skip}, an unsupervised model for sentence embeddings that is trained to predict sentences surrounding a source sentence from its representation. Skip-Thoughts consists of an encoder and two decoders, all of which are parameterized as Gated Recurrent Units (GRUs) \citep{cho2014learning}. The encoder produces a fixed-size representation $f_\theta(s_i)$ for $s_i \in \mathcal{S}_{\mathcal{Z}}$, and the two decoders reconstruct the previous sentence $s_{i-1}$ and the next sentence $s_{i+1}$ in an identical manner, as follows:
\begin{align*}
\mathbf{h}_t &= \text{GRU}(w_{<t}, f_\theta(s_i)),\\
P(w_t | w_{<t}, f_\theta(s_i)) &\propto \exp(\mathbf{v}_{w_t}^T \mathbf{h}_t),
\end{align*}
where $(w_1, \ldots, w_T)$ is the sequence of words in $s_{i-1}$ or $s_{i+1}$ and $\mathbf{v}_w$ denotes an output embedding for $w$. The loss minimizes $- \sum_t \log{P(w_t | w_{<t}, f_\theta(s_i))}$.

In order to learn hyperbolic embeddings, the loss must depend directly on the hyperbolic distance between the source and target embeddings. As an intermediate step, we present a modified version of Skip-Thoughts where we remove the GRU from the decoding step and instead directly predict a bag-of-words surrounding the source sentence, as follows:
\begin{align*}
\mathbf{c}_t =& \frac{1}{2K} \sum_{k=1}^{K} \mathbf{v}'_{w_{t-k}} + \mathbf{v}'_{w_{t+k}},\\
P(w_t | w_{\neq t}, f_\theta(s_i)) \propto& \exp{(\mathbf{v}_{w_t}^T f_\theta(s_i) + \mathbf{v}_{w_t}^T \mathbf{c}_t)}.
\end{align*}
Here, $\mathbf{c}_t$ is an average word embedding of the bi-directional local context around the word to be predicted. We found it was important to condition the prediction on $\mathbf{c}_t$ in order to learn a good quality encoder model $f_\theta$, since it can take care of uninteresting language modeling effects. Empirically, the sentence encoder trained in this manner gives around $1\%$ lower average performance on downstream tasks (discussed in Section \ref{sec:exp_sentences}) than the original Skip-Thoughts model, while being considerably faster. More importantly, the prediction probability now directly depends on the inner product between $\mathbf{v}_{w_t}$ and $f_\theta(s_i)$. We can now introduce a hyperbolic version of the likelihood as follows:
\begin{align*}
&P(w_t | w_{\neq t}, f_\theta(s_i)) \propto\\
&\exp{\left(-\lambda_1 d(\mathbf{v}_{w_t}, f_\theta(s_i)) - \lambda_2 d(\mathbf{v}_{w_t}, \mathbf{c}_t)\right)}.
\end{align*}
Here, $d$ is the hyperbolic distance function (Equation~\ref{eq:hyperbolic}) and $\lambda_1, \lambda_2$ are learned coefficients that control the importance of the two terms. After training, we observed that $\lambda_2 > \lambda_1$, which supports our intuition that local context is more important in predicting a word. 

To ensure that $\mathbf{v}_{w_t}, f_\theta(s_i), \mathbf{c}_t \in \mathcal{B}^d$, we use the following parameterization:
\begin{align*}
\phi_{dir}(\mathbf{x}) = W_1^T \mathbf{x}, \quad \phi_{norm}(\mathbf{x}) = W_2^T \mathbf{x},
\end{align*}
where $\mathbf{x}=\{\mathbf{\hat{v}}_{w_t}, \mathbf{\hat{c}}_t, \hat{f}_\theta(s_i)\}$; $\mathbf{\hat{v}}_{w_t}$ is the Euclidean output embedding for word $w_t$, obtained from a lookup table; $\mathbf{\hat{c}}_t$ is the Euclidean local context embedding, obtained by averaging Euclidean word vectors from a window around $w_t$; and $\hat{f}_\theta$ is a bidirectional GRU encoder over the words of $s_i$:
\begin{align*}
\mathbf{h}^f_T = \overrightarrow{\text{GRU}}(s_i)&, \quad \mathbf{h}^b_1 = \overleftarrow{\text{GRU}}(s_i)\\
\hat{f}_\theta(s_i) &= \mathbf{h}^f_T \| \mathbf{h}^b_1
\end{align*}

Similar to Skip-Thoughts, the loss minimizes  $- \sum_t \log{P(w_t | w_{\neq t}, f_\theta(s_i))}$.

\section{Experiments \& Results}

\subsection{WordNet}


The WordNet noun hierarchy is a collection of tuples $\mathcal{D} = \{(u,v)\}$, where each pair $(u,v)$ denotes that $u$ is a hypernym of $v$. Following \citet{nickel2017poincar}, we learned embeddings using the transitive closure $\mathcal{D}^{+}$, which consists of 82,114 nouns and 743,241 hypernym-hyponym edges. 
We compared our results to the original method from \citet{nickel2017poincar} across three different embedding sizes. 
In each case, we evaluated the embeddings by attempting to reconstruct the WordNet tree using the nearest neighbors of the nodes. For each node, we retrieved a ranked list of its nearest neighbors in embedding space and computed the \textit{mean rank} of its ground truth children, and also computed the \textit{Mean Average Precision} (MAP), which is the average precision at the threshold of each correctly retrieved child. Results are presented in Table~\ref{tab:wordnet}.

\begin{table}[t!]
\scriptsize
\centering
\begin{tabular}{@{}lcccc@{}}
\toprule
\multirow{2}{*}{\textbf{Method}}             &
& \multicolumn{3}{c}{\textbf{Dim}} \\ \cmidrule(l){3-5} 
                                    &                         & \textbf{5}       & \textbf{20}       & \textbf{100}      \\ \midrule
\multicolumn{5}{l}{\textbf{From \citet{nickel2017poincar}}}                                                      \\ \midrule
\multirow{2}{*}{Euclidean}          & Rank               &    3542.3     &     1685.9     &   1187.3       \\
                                    & MAP                     &     0.024    &   0.087       &  0.162        \\ \midrule
\multirow{2}{*}{Poincar\'e}           & Rank               &    4.9     &     3.8     &   3.9       \\
                                    & MAP                     &   0.823      &     0.855     &  0.857        \\ \midrule
\multicolumn{5}{l}{\textbf{This work}}                                                                      \\ \midrule
\multirow{2}{*}{Poincar\'e (re-parameterized)} & Rank               &    10.7     &   6.3       &     5.5     \\
                                    & MAP                     &    0.736     &     0.875     &  0.818        \\ \bottomrule
\end{tabular}
\caption{Reconstruction errors for various embedding dimensions on WordNet.}
\label{tab:wordnet}
\end{table}

\begin{figure}[t!]
\centering
\includegraphics[width=0.8\linewidth]{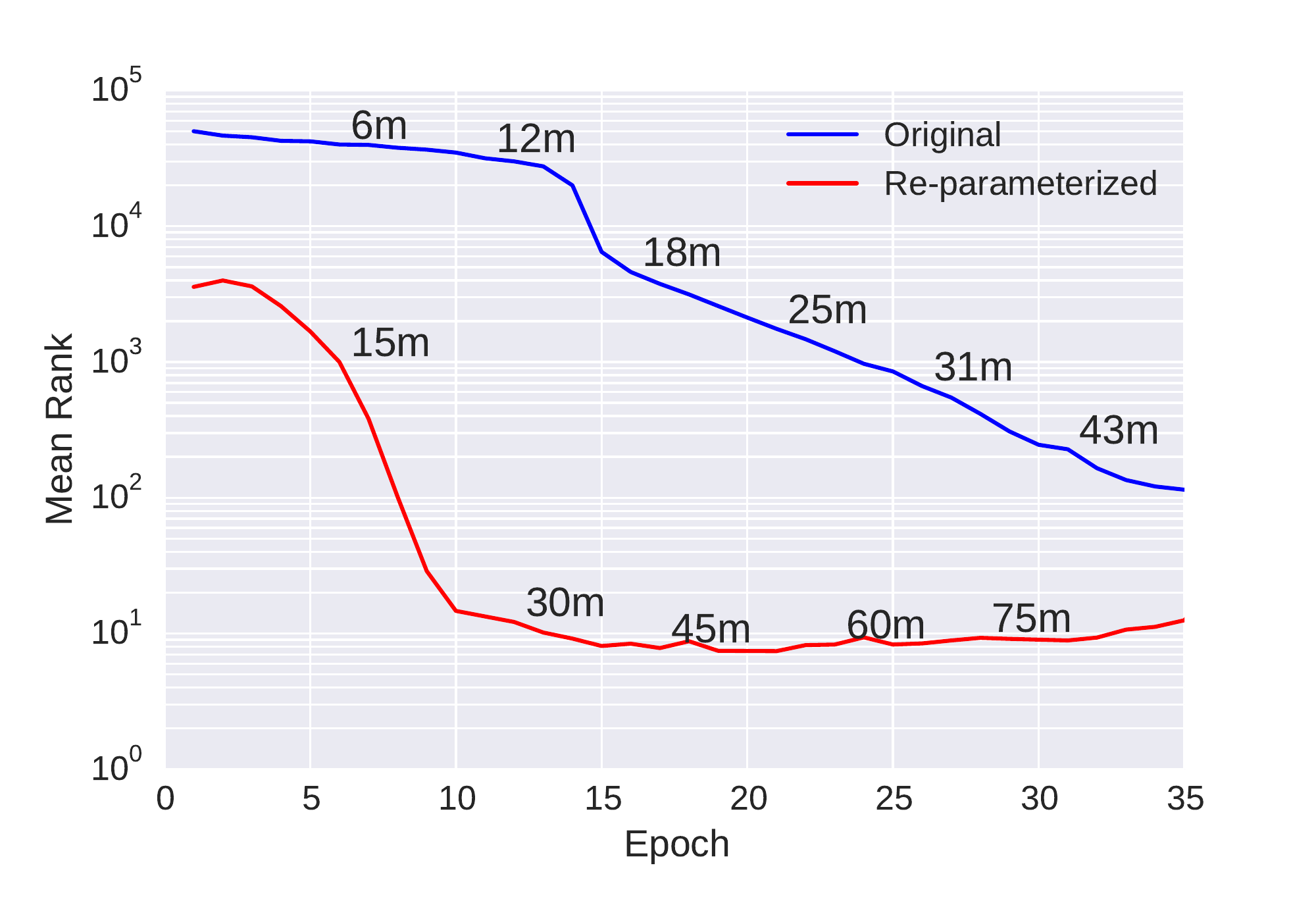}
\caption{Mean Rank for reconstructing the WordNet graph after each training epoch (up to epoch 35) for the original Poincar\'e embeddings method \citep{nickel2017poincar} and our re-parameterized version. Wall time elapsed in minutes is also shown against the curves. Dimension $d=10$.}
\label{fig:timing}
\end{figure}

The re-parameterized Poincar\'e embeddings method has comparable reconstruction error to the original Poincar\'e method, whereas both are significantly superior to the Euclidean embeddings method. Figure \ref{fig:timing} shows reconstruction error after each epoch when training the original and re-parameterized Poincar\'e embeddings, along with the elapsed wall time in minutes\footnote{For the original method we used the official code release at \url{https://github.com/facebookresearch/poincare-embeddings} with the recommended hyperparameter settings. Our re-parameterized model is implemented in TensorFlow \citep{abadi2016tensorflow}. Wall time was recorded on a CPU with 8-core AMD Opteron 6376 Processor.}. The re-parameterized method converges much faster, with its best error achieved around epoch 20, compared to the original method that reaches its best error after hundreds of epochs. This is despite using a larger batch size of 1024 for the re-parameterized method than the original method, which uses batch size 50. We hypothesize that the speed-up is largely due to using the Adam optimizer, which is made possible by the fact that the re-parameterization ensures the embeddings always lie within the Poincar\'e ball.


\subsection{Word Embeddings}

We used the \textsc{text8} corpus\footnote{\url{http://mattmahoney.net/dc/text8.zip}} to evaluate our technique for learning non-parametric unsupervised word embeddings (Section~\ref{sec:nonparam_unsup}). Though small ($17$M tokens), the \textsc{text8} corpus is a useful benchmark for quickly comparing embedding methods.

\begin{table}[t!]
\scriptsize
\centering
\begin{tabular}{@{}lc@{}}
\toprule
\textbf{Word} & \textbf{Nearest neighbors}                                                                            \\ \midrule
vapor         & boiling, melting, evaporation, cooling, vapour             \\ \midrule
towering      & eruptions, tsunamis, hotspots, himalayas, volcanic         \\ \midrule
mercedes      & dmg, benz, porsche, clk, mclaren                           \\ \midrule
forties       & twenties, thirties, roaring, koniuchy, inhabitant          \\ \midrule
eruption      & caldera, vents, calderas, limnic, volcano                  \\ \midrule
palladium     & boron, anion, uranium, ceric, hexafluoride                 \\ \midrule
employment    & incentives, benefits, financial, incentive, investment      \\ \midrule
weighed       & tonnes, weigh, weighs, kilograms, weighing                 \\ \bottomrule
\end{tabular}
\caption{Nearest Neighbors in terms of cosine distance for Poincar\'e embeddings of words ($d=20$).}
\label{tab:nearest-neighbors}
\end{table}

For hyperbolic embeddings, the nearest neighbors of most words by hyperbolic distance (Equation~\ref{eq:hyperbolic}) are all uninteresting common words (e.g. numbers, quantifiers, etc), because points near the origin are relatively close to all points, whereas distances between points increases exponentially as the points approach the boundary of $\mathcal{B}^d$. Instead, we find nearest neighbors in hyperbolic space using cosine distance, which is motivated by the fact that the Poincar\'e ball model is conformal: angles between vectors are identical to their Euclidean counterparts. Some nearest neighbors of hyperbolic word embeddings are shown in Table~\ref{tab:nearest-neighbors}. The closest neighbors typically represent one of several semantic relations with the query word. For example, ``boiling'' produces ``vapor'', ``towering'' is a quality of ``eruptions'', ``dmg'' is the parent company of ``mercedes'', ``tonnes'' is a measure of ``weighed'', and so on. This is a consequence of embedding the word-cooccurrence graph, which implicitly represents these relations.

\begin{table*}[!htbp]
\scriptsize
\centering
\begin{tabular}{@{}cccccccccccc@{}}
\toprule
\multicolumn{3}{c}{\textbf{``bank"}}              & \multicolumn{3}{c}{\textbf{``music"}}             & \multicolumn{3}{c}{\textbf{``dog"}}               & \multicolumn{3}{c}{\textbf{``great"}}             \\ \cmidrule(l){1-3} \cmidrule(l){4-6} \cmidrule(l){7-9} \cmidrule(l){10-12}
\textbf{Word} & \textbf{Count} & \textbf{Norm} & \textbf{Word} & \textbf{Count} & \textbf{Norm} & \textbf{Word} & \textbf{Count} & \textbf{Norm} & \textbf{Word} & \textbf{Count} & \textbf{Norm} \\ \cmidrule(l){1-3} \cmidrule(l){4-6} \cmidrule(l){7-9} \cmidrule(l){10-12}
bank          & 1076           & 2.56          & music         & 4470           & 1.58          & dog           & 566            & 3.21          & great         & 4784           & 2.11          \\
bankruptcy    & 106            & 4.61          & musical       & 1265           & 2.56          & dogs          & 184            & 4.27          & greater       & 1502           & 2.51          \\
banking       & 185            & 5.92          & musicians     & 435            & 4.07          & dogme         & 16             & 6.52          & greatest      & 753            & 2.97          \\
bankrupt      & 28             & 5.93          & musician      & 413            & 4.32          & bulldogs      & 8              & 7.08          & greatly       & 530            & 3.46          \\
banks         & 407            & 6.45          & musicals      & 38             & 5.76          & endogenous    & 5              & 7.55          & greatness     & 12             & 6.41          \\
banknote      & 13             & 6.62          & musicology    & 18             & 6.38          & sheepdog      & 5              & 7.73          &               &                &               \\
\bottomrule
\end{tabular}
\caption{Words in order of increasing hyperbolic norm which contain the substring indicated in the top row. Their counts in the \textsc{text8} corpus are also shown. Dimension size $d=20$.}
\label{tab:word-hierarchy}
\end{table*}

Table~\ref{tab:word-hierarchy} shows lists of related words that contain a particular substring in order of increasing hyperbolic norm. We also show the counts in the corpus of these words, which are correlated to the number of contexts they occur in. As expected, words occurring in fewer contexts have higher hyperbolic norm, and this corresponds to increased specificity as we move down the list; for example ``bulldogs'' has a higher norm than ``dog'', and ``greatest'' has a higher norm than ``great''. The Spearman correlation between $1/f$, where $f$ is the frequency of a word in the corpus, and the norm of its embedding is $0.77$.

We quantitatively evaluate hyperbolic embeddings on two tasks against the baseline Skip-Gram with Negative Sampling (SGNS) embeddings \citep{mikolov2013efficient}\footnote{We use the code available at \url{https://github.com/tensorflow/models/tree/master/tutorials/embedding}, which was tuned for the \textsc{text8} corpus.}. The first task is Word-Similarity on the WordSim-353 dataset \citep{finkelstein2001placing}, which measures whether the embeddings preserve semantic similarity between words as judged by humans. We compute Spearman's correlation between ground truth similarity scores and cosine distances in embedding space between all pairs of words in the dataset. The second task is HyperLex \citep{vulic2017hyperlex}, which measures the extent to which embeddings preserve lexical entailment relationships of the form ``X is a type of Y''. These are precisely the kind of relations we hope to capture in the norm of hyperbolic embeddings. Given a pair $(x,y)$ of words, we compute the score for the relationship $\text{is-a}(x,y)$ in the same way as \citet{nickel2017poincar}:
\begin{equation*}
\text{score(is-a}(x,y)\text{)} = -(1+\alpha(\|y\|-\|x\|))d(x,y).
\end{equation*}
If $x$ and $y$ are close and $\|y\|$$<$$\|x\|$, the above score will be positive, implying $x$ is a type of $y$.

\begin{table}[!t]
\scriptsize
\centering
\begin{tabular}{@{}cccccc@{}}
\toprule
\multirow{2}{*}{\textbf{Task}} & \multirow{2}{*}{\textbf{Method}} & \multicolumn{4}{c}{\textbf{Dimension}}                \\ \cmidrule(l){3-6} 
                               &                                  & \textbf{5} & \textbf{20} & \textbf{50} & \textbf{100} \\ \midrule
\multirow{2}{*}{WordSim-353}   & SGNS                             & 0.350      & 0.566       & 0.676       & 0.689        \\
                               & Poincar\'e                       & 0.305      & 0.451       & 0.451       & 0.455        \\ \midrule
\multirow{2}{*}{HyperLex}      & SGNS                             & -0.002     & 0.093       & 0.124       & 0.140        \\
                               & Poincar\'e                       & 0.259      & 0.246       & 0.246       & 0.248        \\ \bottomrule
\end{tabular}
\caption{Spearman's $\rho$ correlation coefficient for Word Similarity and Lexical Entailment tasks using SGNS and Poincar\'e embeddings.}
\label{tab:wordemb}
\end{table}


Table \ref{tab:wordemb} shows the scores on these two tasks for both SGNS and Poincar\'e embeddings for various embedding sizes. SGNS embeddings are superior for preserving word similarities, while Poincar\'e embeddings are superior for preserving lexical entailment. However, the best score for Poincar\'e embeddings is only $0.259$, which is quite low. In comparison to the unsupervised baselines studied in \citet{vulic2017hyperlex}, Poincar\'e embeddings rank second behind the simple Frequency Ratio baseline which, achieves $0.279$\footnote{This does not include baselines that use extra information like WordNet while learning the embeddings.}.

\subsection{Sentence Embeddings}
\label{sec:exp_sentences}

We use the BookCorpus \citep{zhu2015aligning} to learn sentence and phrase embeddings. We pre-process the data into triples of the form $(s_{i-1},s_i,s_{i+1})$ consisting of both full sentences, as in the original Skip-Thoughts model, and sub-sentence sequences of words sampled according to the same lengths as sentences in the corpus. We found that augmenting the dataset in this manner led to consistent improvements on downstream tasks. 

\begin{figure*}[!h]
\small
\centering
\resizebox{.25\linewidth}{!}{%
\begin{tikzpicture}
\Tree [.{4.0}	[.{4.3}	[.{1.9} Most ] 
						[.{3.0}	[.{1.9} of ] 
                        		[.{2.5}	[.{1.7} the ]
                                		[.{2.2} 10 ] ] ] ]
                [.{2.9}	[.{1.6} have ]
                		[.{3.3}	[.{1.6} big ]
                        		[.{1.7} UNK ]
                                [.{2.5} operations ] ] ] ]
\end{tikzpicture}
}\qquad
\resizebox{.25\linewidth}{!}{%
\begin{tikzpicture}
\Tree [.{2.5}	[.{1.7} No ]
				[.{1.8} , ]
                [.{1.5} it ]
                [.{2.4}	[.{2.5} was ]
                		[.{2.4} n't ]
                        [.{2.2}	[.{2.8} Black ]
                        		[.{1.9} Monday ] ] ] ]
\end{tikzpicture}
}\qquad
\resizebox{.2\linewidth}{!}{%
\begin{tikzpicture}
\Tree [.{7.7}	[.{1.5} It ]
				[.{2.7}	[.{1.9} would ]
                		[.{2.2}	[.{2.0} open ]
                        		[.{2.0}	[.{1.8}	[.{1.7} a ]
                                				[.{1.8} can ] ]
                                        [.{2.4}	[.{1.9} of ]
                                        		[.{2.1} worms ] ] ] ] ] ]
\end{tikzpicture}
}
\caption{Constituent parse trees from the Penn Treebank with hyperbolic norms of the phrase embeddings at each node.}
\label{fig:parsetrees}
\end{figure*}
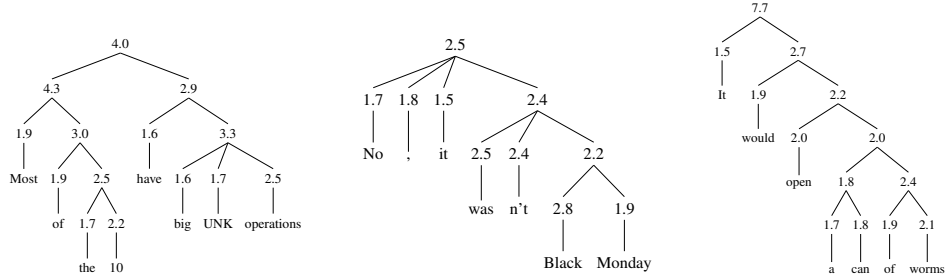

\begin{table}[!t]
\scriptsize
\centering
\begin{tabular}{@{}lc@{}}
\toprule
\textbf{Sentence}                                                                                                                 & \textbf{Norm} \\ \midrule
a creaky staircase gothic .                                                                                                       & 6.21          \\
it 's a rare window on an artistic collaboration .                                                                                & 6.32          \\
a dopey movie clothed in excess layers of hipness .                                                                               & 6.35          \\
an imponderably stilted and self-consciously arty movie .                                                                         & 6.65          \\
there's a delightfully quirky movie ... , but brooms isn't it .                                                  & 6.83          \\
a trifle of a movie, with a few laughs ... unremarkable soft center .                                           & 6.86          \\
\bottomrule
\end{tabular}
\caption{Sentences from Movie Reviews dataset with their norms. Each row represents a nearest neighbor to and with a greater norm than the sentence in the row above.}
\label{tab:sentence-hierarchy}
\end{table}

Similar to word embeddings, we expect that sentence (phrase) embeddings will be organized in a hierarchical manner such that sentences (phrases) that appear in a variety of contexts are closer to the origin. However, unlike word embeddings where we could compare hyperbolic norm to frequency in the corpus, this effect is hard to measure directly for sentences (phrases) because most only appear a small number of times in the corpus. Instead, we check whether the embeddings exhibit a known hierarchical structure: constituent parses of sentences. We take Section $23$ from the Wall Street Journal subset of Penn Treebank \citep{marcus1993building}, which is annotated with gold standard constituent parse tree structures, and embed each node from each tree using the learned parametric encoder $f_\theta$. The Spearman correlation between the norm of the resulting embedding and the height of the node in its tree, computed over all nodes in the set, is $0.671$. Figure~\ref{fig:parsetrees} shows some example parses with the hyperbolic norm at each node. The norms generally increase as we move upwards, indicating that the learned embeddings encode some of this particular form of hierarchical structure. Table~\ref{tab:sentence-hierarchy} shows examples from the Movie Review corpus \citep{pang2002thumbs}, in which we generated a chain of sentences with increasing norm by iteratively searching for the nearest neighbor with norm greater than the previous sentence.


\begin{table*}[!htbp]
\scriptsize
\centering
\begin{tabular}{@{}cccccccccc@{}}
\toprule
\textbf{Encoder Dim}  & \textbf{Word Dim}    & \textbf{Method} & \textbf{Perplexity} & \textbf{CR} & \textbf{SUBJ} & \textbf{MPQA} & \textbf{MR} & \textbf{MultiNLI} & \textbf{SNLI} 
\\ \midrule
\multirow{2}{*}{10}   & \multirow{2}{*}{100} & Euclidean       & 117                 & 0.639       & 0.582         & 0.689         & 0.546       & 0.419             & 0.483         
\\
                      &                      & Poincar\'e      & 110                 & 0.640       & 0.623         & 0.769         & 0.534       & 0.417             & 0.480         
\\ \midrule
\multirow{2}{*}{100}  & \multirow{2}{*}{200} & Euclidean       & 61                  & 0.719       & 0.882         & 0.823         & 0.694       & 0.534             & 0.692         
\\
                      &                      & Poincar\'e      & 53                  & 0.722       & 0.890         & 0.848         & 0.696       & 0.537             & 0.684         
\\ \midrule
\multirow{2}{*}{1000} & \multirow{2}{*}{620} & Euclidean       & 61                  & 0.804       & 0.925         & 0.860         & 0.742       & 0.617             & 0.741         
\\
                      &                      & Poincar\'e      & 46                  & 0.792       & 0.921         & 0.880         & 0.746       & 0.620             & 0.746         
\\ \midrule
2400                  & 620                  & Skip-Thoughts   & --$^\ast$                  & 0.836       & 0.938         & 0.889         & 0.795       & 0.650             & 0.766         
\\ \bottomrule
\end{tabular}
\caption{Held out set perplexity and downstream task performance for sentence embeddings of various sizes. $^\ast$Perplexity of the Skip-Thoughts model is not comparable to our methods since it only uses uni-directional local context.}
\label{tab:skipthoughts}
\end{table*}


Next, following common practice for evaluating sentence representations, we evaluate the trained Poincar\'e encoder as a black-box feature extractor for downstream tasks. We choose four binary classification benchmarks from the original Skip-Thoughts evaluation -- CR, MR, MPQA and SUBJ -- and two entailment tasks -- MultiNLI \citep{williams2017broad} and SNLI \citep{bowman2015large}. For the binary classification tasks we train SVM models with a kernel based on hyperbolic distance between sentences, and for the entailment tasks we train multi-layer perceptrons on top of the premise and hypothesis embeddings and their element-wise products and differences. As a baseline, we compare to embeddings trained using the Euclidean distance metric. Table~\ref{tab:skipthoughts} reports the results of these evaluations for various embedding dimensions. Poincar\'e embeddings achieve a lower perplexity in each case, suggesting a more efficient use of the embedding space. However, both sets of embeddings perform similarly on downstream tasks, except for the MPQA opinion polarity task where Poincar\'e embeddings do significantly better. Training with embedding sizes greater than $1000$ did not show any further improvements in our experiments.


\subsection{Discussion}
The goal of this work was to explore whether hyperbolic spaces are useful for learning embeddings of natural language data.
Ultimately, the usefulness of an embedding method depends on its performance on downstream tasks of interest. In that respect we found mixed results in our evaluation.
For word embeddings, we found that hyperbolic embeddings preserve co-occurrence frequency information in their norms, and this leads to improved performance on a lexical entailment task. However, decreased performance on a word similarity task means that these embeddings may not be useful across all tasks. In general, this suggests that different architectures are needed for capturing different types of lexical relations. We experimented with several other loss functions, pre-processing techniques and hyper-parameter settings, which we did not describe in this paper due to space constraints, but the conclusions remained the same. 

For sentence embeddings, we found evidence that hyperbolic embeddings preserve phrase constituency information in their norms. A deeper investigation of the learned hierarchy is difficult since our encoder is a parametric function over a (practically) infinite set and there is no clear notion of edges in the learned embeddings. On downstream tasks, we saw a small improvement over the Euclidean baseline in some cases and a small degradation in others, again highlighting the need for specialized embeddings for different tasks. We hope that our initial study can pave the way for more work on the applicability of the hyperbolic metric for learning useful embeddings of natural language data. 

\section{Related Work}
\citet{tay2017hyperqa} used the hyperbolic distance metric to learn question and answer embeddings on the Poincar\'e ball for question-answer retrieval. The main difference to our work is that we explore unsupervised objectives for learning generic word and sentence representations from a text corpus. Furthermore, we show that by using re-parameterization instead of projection to constrain the embeddings, we can view the distance metric as any other non-linear layer in a deep network and remove the need for Riemannian-SGD. 

Several works have attempted to learn hierarchical word embeddings. 
Order Embeddings \citep{vendrov2015order} and LEAR \citep{vulic2017specialising} are supervised methods that also encode hierarchy information in the norm of the embeddings by adding regularization terms to the loss function. In comparison, our method is unsupervised. HyperVec \citep{nguyen2017hierarchical} is a supervised method which ensures that the hypernymy relation is assigned a higher similarity score in the learned embeddings than other relations such as synonymy. The vector space model for distribution semantics introduced by \citet{henderson2016vector} is unsupervised and re-interprets word2vec embeddings to predict entailment relations between pairs of words.
DIVE \citep{chang2017distributional} is also unsupervised, and achieves a score of $32.6\%$ on the lexical entailment task, but it is unclear how well the embeddings preserve semantic similarity.

For sentence embeddings, several works have looked at improved loss functions for Skip-Thoughts to make the model faster and light-weight \citep{tang2017trimming,tang2017rethinking,tang2017exploring}. \citet{ba2016layer} introduced a layer normalization method that shows consistent improvements when included in the GRU layers in Skip-Thoughts, and we used this in our encoder. More recently, improved sentence representations were obtained using discourse based objectives \citep{jernite2017discourse,nie2017dissent} and using supervision from natural language inference data \citep{conneau2017supervised}.

\section{Conclusion}
We presented a re-parameterization method that allows us to learn Poincar\'e embeddings on top of arbitrary encoder modules using arbitrary distance-based loss functions. We showed that this re-parameterization leads to comparable performance to the original method from \citet{nickel2017poincar} when explicit hierarchical structure is present in the data. When we applied this method to natural language data at the word- and sentence-level, we found evidence of intuitive notions of hierarchy in the learned embeddings. This led to improvements on some -- but not all -- downstream tasks. Future work could either focus on alternative formulations for unsupervised hyperbolic embeddings, or alternative downstream tasks where hierarchical organization may be more useful.

\bibliography{naaclhlt2018}

\begin{thebibliography}{33}
\expandafter\ifx\csname natexlab\endcsname\relax\def\natexlab#1{#1}\fi

\bibitem[{Abadi et~al.(2016)Abadi, Barham, Chen, Chen, Davis, Dean, Devin,
  Ghemawat, Irving, Isard et~al.}]{abadi2016tensorflow}
Mart{\'\i}n Abadi, Paul Barham, Jianmin Chen, Zhifeng Chen, Andy Davis, Jeffrey
  Dean, Matthieu Devin, Sanjay Ghemawat, Geoffrey Irving, Michael Isard, et~al.
  2016.
\newblock Tensorflow: A system for large-scale machine learning.
\newblock In \emph{OSDI}, volume~16, pages 265--283.

\bibitem[{Adcock et~al.(2013)Adcock, Sullivan, and Mahoney}]{adcock2013tree}
Aaron~B Adcock, Blair~D Sullivan, and Michael~W Mahoney. 2013.
\newblock Tree-like structure in large social and information networks.
\newblock In \emph{Data Mining (ICDM), 2013 IEEE 13th International Conference
  on}, pages 1--10. IEEE.

\bibitem[{Ba et~al.(2016)Ba, Kiros, and Hinton}]{ba2016layer}
Jimmy~Lei Ba, Jamie~Ryan Kiros, and Geoffrey~E Hinton. 2016.
\newblock Layer normalization.
\newblock \emph{arXiv preprint arXiv:1607.06450}.

\bibitem[{Bowman et~al.(2015)Bowman, Angeli, Potts, and
  Manning}]{bowman2015large}
Samuel~R. Bowman, Gabor Angeli, Christopher Potts, and Christopher~D. Manning.
  2015.
\newblock A large annotated corpus for learning natural language inference.
\newblock In \emph{Proceedings of the 2015 Conference on Empirical Methods in
  Natural Language Processing (EMNLP)}. Association for Computational
  Linguistics.

\bibitem[{Chamberlain et~al.(2017)Chamberlain, Clough, and
  Deisenroth}]{chamberlain2017neural}
Benjamin~Paul Chamberlain, James Clough, and Marc~Peter Deisenroth. 2017.
\newblock Neural embeddings of graphs in hyperbolic space.
\newblock \emph{arXiv preprint arXiv:1705.10359}.

\bibitem[{Chang et~al.(2017)Chang, Wang, Vilnis, and
  McCallum}]{chang2017distributional}
Haw-Shiuan Chang, ZiYun Wang, Luke Vilnis, and Andrew McCallum. 2017.
\newblock Distributional inclusion vector embedding for unsupervised hypernymy
  detection.
\newblock \emph{arXiv preprint arXiv:1710.00880}.

\bibitem[{Cho et~al.(2014)Cho, van Merri{\"{e}}nboer, G{\"{u}}l{\c c}ehre,
  Bahdanau, Bougares, Schwenk, and Bengio}]{cho2014learning}
Kyunghyun Cho, Bart van Merri{\"{e}}nboer, {\c C}ağlar G{\"{u}}l{\c c}ehre,
  Dzmitry Bahdanau, Fethi Bougares, Holger Schwenk, and Yoshua Bengio. 2014.
\newblock \href {http://www.aclweb.org/anthology/D14-1179} {Learning phrase
  representations using rnn encoder--decoder for statistical machine
  translation}.
\newblock In \emph{Proceedings of the 2014 Conference on Empirical Methods in
  Natural Language Processing (EMNLP)}, pages 1724--1734, Doha, Qatar.
  Association for Computational Linguistics.

\bibitem[{Conneau et~al.(2017)Conneau, Kiela, Schwenk, Barrault, and
  Bordes}]{conneau2017supervised}
Alexis Conneau, Douwe Kiela, Holger Schwenk, Lo{\"i}c Barrault, and Antoine
  Bordes. 2017.
\newblock \href {http://aclweb.org/anthology/D17-1070} {Supervised learning of
  universal sentence representations from natural language inference data}.
\newblock In \emph{Proceedings of the 2017 Conference on Empirical Methods in
  Natural Language Processing}, pages 670--680. Association for Computational
  Linguistics.

\bibitem[{Everaert et~al.(2015)Everaert, Huybregts, Chomsky, Berwick, and
  Bolhuis}]{everaert2015structures}
Martin~BH Everaert, Marinus~AC Huybregts, Noam Chomsky, Robert~C Berwick, and
  Johan~J Bolhuis. 2015.
\newblock Structures, not strings: linguistics as part of the cognitive
  sciences.
\newblock \emph{Trends in cognitive sciences}, 19(12):729--743.

\bibitem[{Finkelstein et~al.(2001)Finkelstein, Gabrilovich, Matias, Rivlin,
  Solan, Wolfman, and Ruppin}]{finkelstein2001placing}
Lev Finkelstein, Evgeniy Gabrilovich, Yossi Matias, Ehud Rivlin, Zach Solan,
  Gadi Wolfman, and Eytan Ruppin. 2001.
\newblock Placing search in context: The concept revisited.
\newblock In \emph{Proceedings of the 10th international conference on World
  Wide Web}, pages 406--414. ACM.

\bibitem[{Henderson and Popa(2016)}]{henderson2016vector}
James Henderson and Diana Popa. 2016.
\newblock \href {https://doi.org/10.18653/v1/P16-1193} {A vector space for
  distributional semantics for entailment}.
\newblock In \emph{Proceedings of the 54th Annual Meeting of the Association
  for Computational Linguistics (Volume 1: Long Papers)}, pages 2052--2062.
  Association for Computational Linguistics.

\bibitem[{Jernite et~al.(2017)Jernite, Bowman, and
  Sontag}]{jernite2017discourse}
Yacine Jernite, Samuel~R Bowman, and David Sontag. 2017.
\newblock Discourse-based objectives for fast unsupervised sentence
  representation learning.
\newblock \emph{arXiv preprint arXiv:1705.00557}.

\bibitem[{Kingma and Ba(2014)}]{kingma2014adam}
Diederik~P Kingma and Jimmy Ba. 2014.
\newblock Adam: A method for stochastic optimization.
\newblock In \emph{Proceedings of the 3rd International Conference on Learning
  Representations (ICLR)}.

\bibitem[{Kiros et~al.(2015)Kiros, Zhu, Salakhutdinov, Zemel, Urtasun,
  Torralba, and Fidler}]{kiros2015skip}
Ryan Kiros, Yukun Zhu, Ruslan~R Salakhutdinov, Richard Zemel, Raquel Urtasun,
  Antonio Torralba, and Sanja Fidler. 2015.
\newblock Skip-thought vectors.
\newblock In \emph{Advances in neural information processing systems}, pages
  3294--3302.

\bibitem[{Krioukov et~al.(2010)Krioukov, Papadopoulos, Kitsak, Vahdat, and
  Bogun{\'a}}]{krioukov2010hyperbolic}
Dmitri Krioukov, Fragkiskos Papadopoulos, Maksim Kitsak, Amin Vahdat, and
  Mari{\'a}n Bogun{\'a}. 2010.
\newblock Hyperbolic geometry of complex networks.
\newblock \emph{Physical Review E}, 82(3):036106.

\bibitem[{Marcus et~al.(1993)Marcus, Marcinkiewicz, and
  Santorini}]{marcus1993building}
Mitchell~P Marcus, Mary~Ann Marcinkiewicz, and Beatrice Santorini. 1993.
\newblock Building a large annotated corpus of english: The penn treebank.
\newblock \emph{Computational linguistics}, 19(2):313--330.

\bibitem[{Mikolov et~al.(2013{\natexlab{a}})Mikolov, Chen, Corrado, and
  Dean}]{mikolov2013efficient}
Tomas Mikolov, Kai Chen, Greg Corrado, and Jeffrey Dean. 2013{\natexlab{a}}.
\newblock Efficient estimation of word representations in vector space.
\newblock \emph{arXiv preprint arXiv:1301.3781}.

\bibitem[{Mikolov et~al.(2013{\natexlab{b}})Mikolov, Sutskever, Chen, Corrado,
  and Dean}]{mikolov2013distributed}
Tomas Mikolov, Ilya Sutskever, Kai Chen, Greg~S Corrado, and Jeff Dean.
  2013{\natexlab{b}}.
\newblock Distributed representations of words and phrases and their
  compositionality.
\newblock In \emph{Advances in neural information processing systems}, pages
  3111--3119.

\bibitem[{Mikolov et~al.(2013{\natexlab{c}})Mikolov, Yih, and
  Zweig}]{mikolov2013linguistic}
Tomas Mikolov, Wen-tau Yih, and Geoffrey Zweig. 2013{\natexlab{c}}.
\newblock Linguistic regularities in continuous space word representations.
\newblock In \emph{Proceedings of the 2013 Conference of the North American
  Chapter of the Association for Computational Linguistics: Human Language
  Technologies}, pages 746--751.

\bibitem[{Nguyen et~al.(2017)Nguyen, K{\"o}per, Schulte~im Walde, and
  Vu}]{nguyen2017hierarchical}
Kim~Anh Nguyen, Maximilian K{\"o}per, Sabine Schulte~im Walde, and Ngoc~Thang
  Vu. 2017.
\newblock \href {http://aclweb.org/anthology/D17-1022} {Hierarchical embeddings
  for hypernymy detection and directionality}.
\newblock In \emph{Proceedings of the 2017 Conference on Empirical Methods in
  Natural Language Processing}, pages 233--243. Association for Computational
  Linguistics.

\bibitem[{Nickel and Kiela(2017)}]{nickel2017poincar}
Maximillian Nickel and Douwe Kiela. 2017.
\newblock \href
  {http://papers.nips.cc/paper/7213-poincare-embeddings-for-learning-hierarchical-representations.pdf}
  {Poincar\'{e} embeddings for learning hierarchical representations}.
\newblock In I.~Guyon, U.~V. Luxburg, S.~Bengio, H.~Wallach, R.~Fergus,
  S.~Vishwanathan, and R.~Garnett, editors, \emph{Advances in Neural
  Information Processing Systems 30}, pages 6338--6347. Curran Associates, Inc.

\bibitem[{Nie et~al.(2017)Nie, Bennett, and Goodman}]{nie2017dissent}
Allen Nie, Erin~D Bennett, and Noah~D Goodman. 2017.
\newblock Dissent: Sentence representation learning from explicit discourse
  relations.
\newblock \emph{arXiv preprint arXiv:1710.04334}.

\bibitem[{Pang et~al.(2002)Pang, Lee, and Vaithyanathan}]{pang2002thumbs}
Bo~Pang, Lillian Lee, and Shivakumar Vaithyanathan. 2002.
\newblock Thumbs up?: sentiment classification using machine learning
  techniques.
\newblock In \emph{Proceedings of the ACL-02 conference on Empirical methods in
  natural language processing-Volume 10}, pages 79--86. Association for
  Computational Linguistics.

\bibitem[{Pennington et~al.(2014)Pennington, Socher, and
  Manning}]{pennington2014glove}
Jeffrey Pennington, Richard Socher, and Christopher Manning. 2014.
\newblock Glove: Global vectors for word representation.
\newblock In \emph{Proceedings of the 2014 conference on empirical methods in
  natural language processing (EMNLP)}, pages 1532--1543.

\bibitem[{Tang et~al.(2017{\natexlab{a}})Tang, Jin, Fang, Wang, and
  de~Sa}]{tang2017rethinking}
Shuai Tang, Hailin Jin, Chen Fang, Zhaowen Wang, and Virginia de~Sa.
  2017{\natexlab{a}}.
\newblock \href {http://aclweb.org/anthology/W17-2625} {Rethinking
  skip-thought: A neighborhood based approach}.
\newblock In \emph{Proceedings of the 2nd Workshop on Representation Learning
  for NLP}, pages 211--218. Association for Computational Linguistics.

\bibitem[{Tang et~al.(2017{\natexlab{b}})Tang, Jin, Fang, Wang, and
  de~Sa}]{tang2017exploring}
Shuai Tang, Hailin Jin, Chen Fang, Zhaowen Wang, and Virginia~R de~Sa.
  2017{\natexlab{b}}.
\newblock Exploring asymmetric encoder-decoder structure for context-based
  sentence representation learning.
\newblock \emph{arXiv preprint arXiv:1710.10380}.

\bibitem[{Tang et~al.(2017{\natexlab{c}})Tang, Jin, Fang, Wang, and
  de~Sa}]{tang2017trimming}
Shuai Tang, Hailin Jin, Chen Fang, Zhaowen Wang, and Virginia~R de~Sa.
  2017{\natexlab{c}}.
\newblock Trimming and improving skip-thought vectors.
\newblock \emph{arXiv preprint arXiv:1706.03148}.

\bibitem[{Tay et~al.(2018)Tay, Tuan, and Hui}]{tay2017hyperqa}
Yi~Tay, Luu~Anh Tuan, and Siu~Cheung Hui. 2018.
\newblock \href {https://doi.org/10.1145/3159652.3159664} {Hyperbolic
  representation learning for fast and efficient neural question answering}.
\newblock In \emph{Proceedings of the Eleventh ACM International Conference on
  Web Search and Data Mining}, WSDM '18, pages 583--591, New York, NY, USA.
  ACM.

\bibitem[{Vendrov et~al.(2015)Vendrov, Kiros, Fidler, and
  Urtasun}]{vendrov2015order}
Ivan Vendrov, Ryan Kiros, Sanja Fidler, and Raquel Urtasun. 2015.
\newblock Order-embeddings of images and language.
\newblock \emph{arXiv preprint arXiv:1511.06361}.

\bibitem[{Vuli{\'c} et~al.(2017)Vuli{\'c}, Gerz, Kiela, Hill, and
  Korhonen}]{vulic2017hyperlex}
Ivan Vuli{\'c}, Daniela Gerz, Douwe Kiela, Felix Hill, and Anna Korhonen. 2017.
\newblock Hyperlex: A large-scale evaluation of graded lexical entailment.
\newblock \emph{Computational Linguistics}, 43(4):781--835.

\bibitem[{Vuli{\'c} and Mrk{\v{s}}i{\'c}(2017)}]{vulic2017specialising}
Ivan Vuli{\'c} and Nikola Mrk{\v{s}}i{\'c}. 2017.
\newblock Specialising word vectors for lexical entailment.
\newblock \emph{arXiv preprint arXiv:1710.06371}.

\bibitem[{Williams et~al.(2017)Williams, Nangia, and
  Bowman}]{williams2017broad}
Adina Williams, Nikita Nangia, and Samuel~R Bowman. 2017.
\newblock A broad-coverage challenge corpus for sentence understanding through
  inference.
\newblock \emph{arXiv preprint arXiv:1704.05426}.

\bibitem[{Zhu et~al.(2015)Zhu, Kiros, Zemel, Salakhutdinov, Urtasun, Torralba,
  and Fidler}]{zhu2015aligning}
Yukun Zhu, Ryan Kiros, Rich Zemel, Ruslan Salakhutdinov, Raquel Urtasun,
  Antonio Torralba, and Sanja Fidler. 2015.
\newblock Aligning books and movies: Towards story-like visual explanations by
  watching movies and reading books.
\newblock In \emph{Proceedings of the IEEE international conference on computer
  vision}, pages 19--27.

\end{thebibliography}
\bibliographystyle{acl_natbib}

\appendix

\section{Implementation Details}
\subsection{WordNet Experiments}
For our re-parameterized Poincar\'e embeddings we used a batch size $1024$, learning rate $0.005$, and no burn-in period. The loss was optimized using the Adam optimizer. Embeddings were initialized in $\mathcal{U}[-0.001, 0.001]$. We sampled $10$ negatives on the fly during training independently for each positive sample. We clipped gradients to a norm of $5$.
Embeddings were initialized to a small norm around $\sigma(-5)$.

\subsection{Word Embedding Experiments}
The \textsc{text8} corpus contains around $17$M tokens preprocessed such that all tokens are lowercase, numbers are spelled out, and any characters not in a-z are replaced by whitespace. We removed stopwords and constructed the word-cooccurrence graph $\mathcal{G}$ by adding an edge between words appearing within $5$ tokens of each other in the resulting corpus. We used $c=0.25$ for subsampling frequent edges, and trained our embedding model using the Adam optimizer with batch size $512$ and learning rate $0.005$. We sampled $50$ negatives per step for the loss. We initialized the norms of the word embeddings around $\sigma(-5)$. All hyper-parameters were tuned to maximize performance on the word similarity task. 

\subsection{Sentence Embedding Experiments}
During preprocessing, only the top 20,000 most frequent types were retained and the rest were replaced with the UNK type. We optimized the loss function using Adam optimizer with a batch size of $64$. The initial learning rate was tuned between $0.005, 0.0008, 0.0001$ which was then decayed exponentially to half its value in 100,000 steps. When decoding we utilize a local context from a window of $K=2$ words around the target word. The embedding norms are initialized around $\sigma(-2)$.

\end{document}